\documentclass[draft]{agujournal2019}
\usepackage{url} 
\usepackage[inline]{trackchanges} 
\usepackage{soul}
\usepackage[Symbol]{upgreek}

\draftfalse

\journalname{Water Resource Research}

\begin{document}

%
%


\title{Multiphase flow prediction with deep neural networks}

%
%




\authors{Gege Wen\affil{1}, Meng Tang\affil{1}, and Sally M. Benson\affil{1}}

\affiliation{1}{Department of Energy Resource Engineering, Stanford University}




\correspondingauthor{Gege Wen}{gegewen@stanford.edu}





\begin{keypoints}
\item A deep neural network is developed for multiphase flow problems involving the complex interplay of gravity, capillary, and viscous forces.
\item The model provides fast and accurate saturation predictions given heterogeneous permeability maps, injection duration, rate, and location.
\item A transfer learning procedure is introduced that can improve a model’s performance when extrapolating without massive data collection.
\end{keypoints}
%
%

%
%


\begin{abstract}
This paper proposes a deep neural network approach for predicting multiphase flow in heterogeneous domains with high computational efficiency. The deep neural network model is able to handle permeability heterogeneity in high dimensional systems, and can learn the interplay of viscous, gravity, and capillary forces from small data sets. Using the example of carbon dioxide (CO$_2$) storage, we demonstrate that the model can generate highly accurate predictions of a CO$_2$ saturation distribution given a permeability field, injection duration, injection rate, and injection location. The trained neural network model has an excellent ability to interpolate and to a limited extent, the ability to extrapolate beyond the training data ranges. To improve the prediction accuracy when the neural network model needs to extrapolate, we propose a transfer learning (fine-tuning) procedure that can quickly teach the neural network model new information without going through massive data collection and retraining. Based on this trained neural network model, a web-based tool is provided that allows users to perform CO${_2}$-water multiphase flow calculations online. With the tools provided in this paper, the deep neural network approach can provide a computationally efficient substitute for repetitive forward  multiphase flow simulations, which can be adopted to the context of history matching and uncertainty quantification. 
\end{abstract}


\section{Background}

Multiphase flow in porous media is an important process for a wide range of engineering problems including nuclear waste management, groundwater contamination, oil and gas production, and carbon capture and storage. Numerical simulation is an essential tool for quantifying these processes by solving spatially and temporally discretized mass and energy balance equations. However, due to the inherent uncertainty in the spatial distribution of geological attributes of the subsurface, a single deterministic model is not sufficient for predicting performance for most multiphase flow problems. In practice, to increase the confidence of simulation results and aid engineering decisions, methods of inverse modeling (also referred as history matching) \cite{Oliver2011} and uncertainty quantification \cite{Tartakovsky2013, Kitanidis2015} have been widely investigated. The goal of inverse modeling is to use the observed reservoir behavior to estimate the reservoir parameters. It often requires a large number of forward simulations, which induce huge computational costs. Uncertainty quantification, on the other hand, evaluates the impact of parameter uncertainty on predicted performance. Conventional uncertainty quantification often rely on Monte Carlo based sampling \cite{robert2013monte}, which also requires large numbers of repetitive forward simulations. 

As an alternative to repetitive full-physics forward simulations, surrogate methods, have been widely studied to improve the computational efficiency for subsurface flow problems (e.g., reduce-order modelling \cite{Cardoso2008, He2013}, polynomial chaos \cite{Bazargan2015}, Gaussian Process \cite{Tian2017, Hamdi2017}, etc). However, these surrogate models generally uses methods that require simplification and reformulation of the problem which may not adequately represent all of the salient features of the full physics models. Meanwhile, the heterogeneous nature of subsurface geology often results in high dimensional inputs. Most intrusive methods suffer from the curse of dimensionality \cite{xiu2010numerical} and fail to work for high-dimensional problems.

To mitigate these challenges for multiphase flow simulation, in this paper, we propose a neural network approach that can achieve accurate and fast multiphase flow prediction. Neural networks are composed of nonlinear nodes and interconnected layers, which can theoretically estimate any complex function given adequate setup and training \cite{Hagan1996, Haykin1999}. In the past decade, the applications of neural networks have exponentially increased due to the development of graphics processing units (GPUs) and data availability \cite{Rusk2015}. Convolutional neural networks (CNN), specifically, are known for their ability to process image inputs. Convolutional layers with filters handle localized spatial features while reducing the number of learnable parameters, making neural networks easy to train \cite{Krizhevsky2012}. In many traditional fields of science and engineering, CNNs are adopted to applications that involve image data \cite{Vandal2018, Reichstein, Shen2018}.

In the field of subsurface flow studies, we have seen some encouraging results that use CNNs in two types of applications. In the first type of applications, CNNs are used for parameterization and probabilistic inversion to generate geological realizations that fit the data while being visually resemble to the real world formation \cite{ Laloy2018, Laloy2017, Liu2018}. Secondly, CNNs can be used to conduct flow predictions, where the mapping from the input properties (e.g. permeability) to the output states (e.g. flow velocity, saturation, pressure) are treated as an image-to-image regression  \cite{Zhu2018, Mo2018, Mo2019, Zhong2019, mo2019b}. Zhu and Zabaras (2018) first proposed an encoder-decoder network to solve a single-phase flow problem where the network predicts the flow velocity and pressure fields given a permeability map. Applying a similar encoder-decoder network, Mo et al. (2019a) studied a CO$_{2}$-water multiphse flow problem, in which the injection duration is added as an additional scalar input to the network and broadcast to a separate channel in the latent space to make predictions over time. This encoder-decoder network structure was also used in a groundwater contaminant transport problem in which the time dependence is captured by an autoregressive model \cite{Mo2019}. A 3D version of this network was published recently where they utilize 3D convolution filters instead of 2D convolution filters to resolve a solute transport problem \cite{mo2019b}. Another approach by Zhong et al. (2019) uses a conditional Generative Adversarial Network to conduct the input-output mapping. Their study also focuses on a CO$_{2}$-water multiphse flow problem and the injection duration is included as a conditional number that serves the training as auxiliary information. In the aforementioned 2D studies, the permeability fields all lie on the $xy$ plane, and therefore the gravity force is not considered.

Building on previous work, here we take an essential step towards adopting neural network based surrogate models in practical settings by embedding spatial information in the controlling parameters. With this functionality, we can control not only the duration and rate of the injection, but also the location where we inject. The system that we are interested in lies in the $rz$ plane, therefore viscous, capillary, and gravity forces all play an important role. Our test cases demonstrate that the neural network proposed here can achieve very high prediction accuracy in the CO${_2}$-water multiphase flow system with complex and discontinuous geological heterogeneities.

In this paper, we consider a carbon capture and storage (CCS) example to demonstrate the potential of using deep neural networks for multiphase flow problems. CCS is an essential greenhouse gas mitigation technology for achieving the 2 degrees C target \cite{IPCC2014}. Carbon dioxide (CO$_{2}$) captured from concentrated sources, the atmosphere, or through bio-energy production, is compressed into a liquid and injected into deep geological formations for long term sequestration. Once CO$_{2}$ is injected into the formation, it migrates away from the injection well while rising upward since CO$_{2}$ is lighter than the formation fluid. A top seal composed of low permeability rock such as shale retains the CO$_{2}$ in the storage reservoir. In the event of encountering faults, fractures, or well infrastructures, the injected CO$_{2}$ is subjected to the risk of leakage. As a result, fast and accurate simulations of CO$_{2}$ migration in complex geological settings is a prerequisite for effective CCS projects~\cite{U.S.DepartmentofEnergy2017}. 

The body of this paper is organized as follows. Section 2 introduces the problem statement, which includes the governing equations and numerical simulation setup. Section 3 describes the methodology of using deep neural networks for multiphase flow prediction, which includes data set preparation, model training, and model prediction. The results are discussed in Section 4, along with a sensitivity study on the size of training data set. In section 5, we address the key challenge in neural network prediction, which is the the ability to generalize outside of the training data range, and provide the method of transfer learning to improve the model's predictive capability in these circumstances. 

\section{Problem statement}
Our focus in this paper is on a CO$_2$ storage problem, where the multiphase flow is composed by CO$_2$ and water in supercritical and liquid phase. We are interested in a 2-d radially symmetrical system which replicates the geometry of CO$_2$ being injected into a storage reservoir through an injection well. In this section, we first introduce the governing equations of this CO$_2$-water multiphase flow problem, then discuss the numerical simulation approach to this problem.


\subsection{Governing equations}
For multiple-phase flow, the basic mass- and energy balance equations is described as \cite{Pruess1999}

\begin{equation}
\frac{\mathrm{d} }{\mathrm{d} t}\int_{V_n}M^\kappa dV_n = \int_{\Upgamma}\mathbf{F}^\kappa \cdot \mathbf{n} d\Upgamma_n + \int_{V_n}q^\kappa dV_n
\end{equation}
over an arbitrary sub-domain of $V_n$ bounded by a closed surface $n$. $M$ denotes the mass accumulation term, $\mathbf{F}$ is the mass flux, $q$ is the source term, and $\mathbf{n}$ is a normal vector on $d\Upgamma_n$ pointing into the $V_n$. In this paper, we are interested in the problem where the components $\kappa$ are $CO_2$ and water. For each component, we have the mass conservation described as
\begin{equation}
\frac{\partial M^\kappa}{\partial t}=-\nabla\cdot \mathbf{F}^\kappa+q^\kappa.
\end{equation}
The mass accumulation term is summed over phases $p$
\begin{equation}
M^\kappa=\phi\sum_pS_p\rho_pX^\kappa_p,
\end{equation}
where $\phi$ is the porosity, $S_p$ is the saturation of phase $p$, $\rho_p$ is the density of phase $p$, and $X^\kappa_p$ is the mass fraction of component $\kappa$ present in phase $p$. 

For each component $\kappa$, we also have the advective mass flux $\mathbf{F^\kappa}$ obtained by summing over phases $p$,
\begin{equation}
\mathbf{F}^\kappa|_{adv}=\sum_{p}X^\kappa_p\mathbf{F}_{p}
\end{equation}
where each individual phase flux $\mathbf{F}_{p}$ is governed by Darcy's law:
\begin{equation}
\mathbf{F}_p = \rho_p \mathbf{u}_p = -k\frac{k_{rp}\rho_p}{\mu_p}(\nabla P_{p} - \rho_p\mathbf{g}).
\end{equation}
Here $\mathbf{u}_p$ is the Darcy velocity of phase $p$, $k$ is the absolute permeability, $k_{rp}$ is the relative permeability of phase $p$, and $\mu_p$ is the viscosity of phase $p$. The fluid pressure of phase $p$
\begin{equation}
P_p = P + P_{cp}
\end{equation}
is the sum of the reference phase (usually the gas phase) pressure $P$ and the capillary pressure $P_{cp}$. Here $\mathbf{g}$ is the gravitational acceleration. For the scope of this paper, our simulation does not include molecular diffusion and hydrodynamic dispersion.

\subsection{Numerical simulation setup}
The numerical simulation of the above problem is performed by a state-of-the-art full-physics numerical simulator ECLIPSE. ECLIPSE uses upstream weighting and the Adaptive IMplicit method for simulation. \cite{Schlumberger}

In this study, we are interested in a CO$_2$ plume migration problem that lies in the $rz$ plane. This a challenging problem for numerical simulation due to the presence of the strong non-linearity. In particular, as compared to a $xy$ problem, gravity plays an important role in this problem and the permeability heterogeneity, especially layered heterogeneity, can significantly affect the migration of the CO$_2$ plume \cite{gege2019}. 

The simulated column is a 2-d radially symmetrical system with an injection well located in the center of the cylindrical volume. The top and bottom boundary conditions are no-flow boundaries and the horizontal boundary is a constant pressure boundary. A uniform grid of $128 \times 128$ with a grid size of 1 m is used in the numerical simulation. Each simulation takes around 15 minutes on an Intel Core i7-4790 CPU. Relevant parameters used for the simulations are summarized in Table \ref{simulation_param}.

\begin{table}[!htp]
 \caption{ECLIPSE Simulation Parameters}
  \centering
  \begin{tabular}{ll}
    \hline
    \textbf{Parameter}     &  \textbf{Value}     \\
    \hline
    Formation Depth    &   1000m                  \\
    Porosity &  0.2  \\
    Temperature (isothermal) &  $40^\circ$C \\
    Salinity &  0 \\
     \hline
    \textbf{Capillary pressure curve} - van Genuchten  \\
     \hline
    $\lambda$ & 0.3 \\
    S$_{ls}$ & 0.999   \\
    P$_{max}$ & 2.1E7 Pa   \\
     \hline
    \textbf{Capillary entry pressure} - Leverett J-function  \\
     \hline
    k$_{ref}$ &  3.95E-14 m$^2$   \\
    $\phi_{ref}$ &  0.185   \\
    P$_{ref}$ &  7.5E3 Pa   \\
     \hline
    \textbf{Relative permeability} - Corey's curve  \\
     \hline
    S$_{lr}$ & 0.3    \\
    S$_{gr}$ & 0    \\
    \hline
  \end{tabular}
  \label{simulation_param}
\end{table}

\section{Methodology}
\begin{figure}[!htp]
  \centering
  \includegraphics[scale = 0.3]{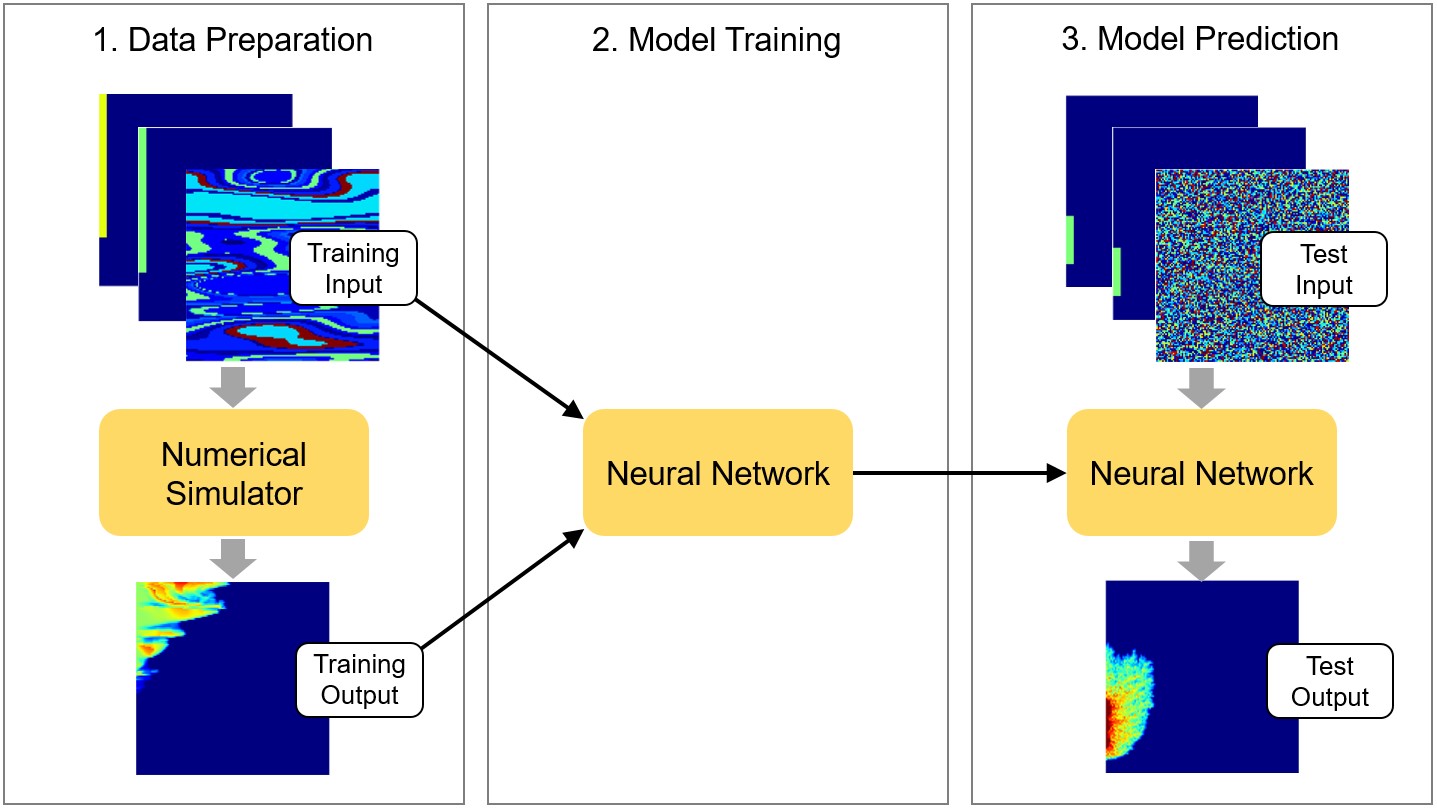}
  \caption{The work flow of using neural networks to predict multiphase flow results.}
  \label{fig:workflow}
\end{figure}

This section introduces the procedures of using a deep neural network to perform prediction for a multiphase flow problem, which includes data set preparation; model training; and model prediction. The workflow is illustrated in Figure \ref{fig:workflow} and each stage is discussed in a subsection. The data set preparation section introduces the mapping between the training input and the training output, as well as the design of a representative data set. In the model training section, we introduce our neural network architecture; the training loss, as well as the training process. The model prediction section discusses the model's test time specifications and evaluation metrics.

\subsection{Data set preparation} 

\begin{figure}[!htp]
  \centering
  \includegraphics[scale = 0.37]{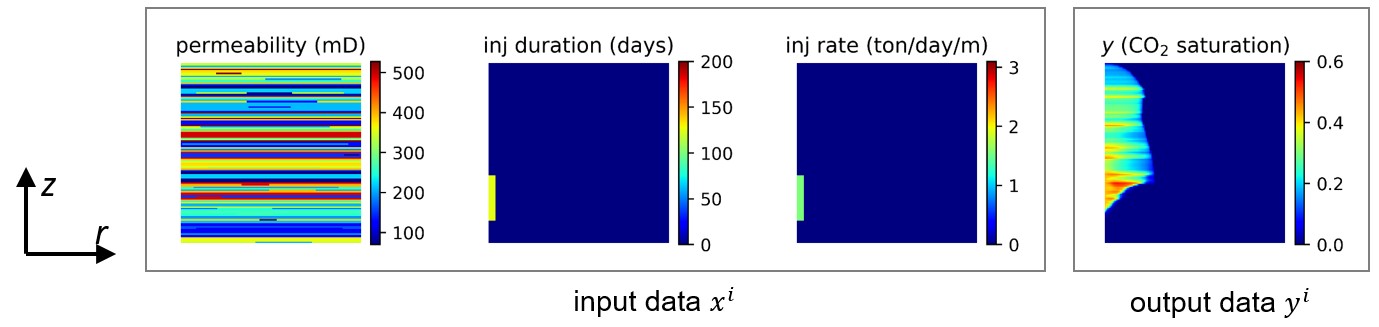}
  \caption{An example from the training set. The permeability, injection duration, injection rate, and saturation fields all lie in a $rz$ plane, which can be seen as cross sections along the radius. The modeled volume is radially symmetrical and the injection well is located in the center of the cylindrical volume.}
  \label{data_example}
\end{figure}

In order to the train the neural network to predict the CO$_2$ plume migration given the formation geology as well as the injection information, the data set has to contain mappings from the various combinations of a permeability field, an injection duration field, and an injection rate field to the corresponding CO$_2$ saturation plume field. Figure \ref{data_example} shows an data example of this mapping.

\begin{figure}[!htp]
  \centering
  \includegraphics[scale = 0.6]{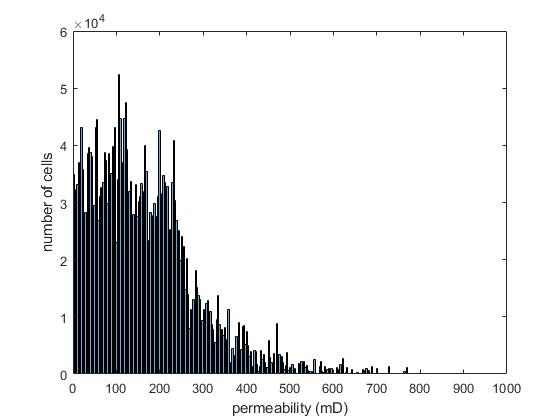}
  \caption{Permeability histogram of the 200 randomly generated fields being used in the training and test set.}
  \label{histogram}
\end{figure}

To mimic the sedimentary formations that are typically used for sequestration, 200 permeability fields are randomly generated using the Stanford Geostatistical Modeling Software (SGeMS)\cite{SGeMS} with laterally correlated heterogeneity. Correlation lengths for the permeability heterogeneity distributions range from a few meters to several hundred meters and input parameters of SGeMs is summarized in Table\ref{sgems_param}. The formation heterogeneity is radially symmetrical with an injection well located in the center cylindrical volume as shown in Figure \ref{data_example}. We do not claim these are realistic models of any reservoir, but use them as a way to demonstrate the neural network's predictive ability in highly heterogeneous systems. Note that the permeability fields studied here are non-Gaussian and contains discontinuous heterogeneity in high dimensional spaces, which is very challenging for most surrogate methods. The histogram of permeability on each grid for the 200 fields are shown in Figure \ref{histogram}. 

\begin{table}[!htp]
 \caption{SGeMS input parameters for random field generation}
  \centering
  \begin{tabular}{lll}
    \hline
    Parameter   & unit  & Value     \\
    \hline
    lateral heterogeneity correlation length (ax) & m   &  random integer in the interval (1, 256)    \\
    vertical heterogeneity correlation length (az) & m &  random integer in the interval (1, ax)  \\
    permeability mean in each field  & mD& random number in the interval (1, 350) \\
    permeability standard deviation in each field  & mD & random number in the interval (1, 200) \\
    \# of materials in each field & - & 20 \\
    fraction assigned to each material & - & random number in the interval (0, 1) \\
    \hline
  \end{tabular}
  \label{sgems_param}
\end{table}

Supercritical CO$_{2}$ is injected  with various rates, duration, and well perforation locations. To incorporate these injection information, we create two separate channels which correspond to the injection rate and injection duration respectively. The well perforation is treated as a CO$_{2}$ source term on the designated grid cell location, which is marked as the colored cells in the injection duration and injection rate examples in Figure \ref{data_example}. The injection duration varies from 0 to 200 days and the injection rate varies from 0 to 3.1 ton/day per meters of well perforation. In this example, the injection duration channel shows that the injection lasts for 200 days and the injection rate channel shows that CO$_{2}$ is injected at a rate of 1.6 ton/day per meter of well perforation. The injection well is perforated from 32 m to 96 m and this spatial information is embedded in both channels. 

In order to give the neural network a balanced data set, where the permeability, the injection duration, and injection rate all equally affect the results of CO$_2$ saturation plume, the number of injection duration scenarios, the injection rate scenarios, and the permeability field scenarios needs to be comparable. In our data set, we have 200 randomly generated permeability fields, which constrains the number of injection duration fields and injection rate fields to be comparable to 200. As a result, Our data set contains 36 scenarios of different injection location and 8 injection duration, contributing to a total of $36\times 8 = 288$ scenarios of injection duration fields. Similarly, we have 8 injection rates for each injection location, resulting in a total of $36\times 4 = 144$ scenarios of injection rate fields. In total, the full data set has $200\times 36 \times 8 \times 4 = 230,400$ samples.

In the data set, each permeability field, injection rate field, and injection duration field are concatenated to become one data input. This approach is scalable to more controlling parameters with spatial information (e.g. porosity, initial pressure, initial saturation, etc.), which can be easily stacked to the input. In convolutional neural networks, each of the concatenated field is called a "channel" of the input. During training, the learnable parameters in the convolutional filters will be automatically tuned to handle the information in each channel and this method has proved to be effective even with very small data sets.

\subsection{Model Training}
The following section describes the architecture of the proposed neural network for this multiphase flow prediction problem. We also introduce the training strategy and the choice of loss function in this section.

\subsubsection{Neural network architecture}
The main task of the neural network resembles an image to image regression from the input permeability field, injection duration field, and rate duration field to the output CO$_{2}$ saturation field. To conduct this task efficiently, we introduce the RU-Net, a deep convolutional neural network architecture developed base on the well-known U-Net architecture \cite{Ronneberger2015}, which was originally proposed for bio-medical image segmentation. The U-Net architecture contains an encoding path and a decoding path, which captures the hierarchical spatial encoding and decoding of the input and the output images. The U-Net architecture is especially efficient in data utilization due to the design of the concatenating channel 'highway'. The concatenating channels pass on the multi-scale spatial information obtained from the encoding path to the corresponding decoding path, where the multi-scale encoding can aid the prediction. Between the encoding and the decoding path, a connecting block is designed to learn the relationship between the input and output latent representations. 

\begin{figure}[!htp]
\begin{center}
\centerline{\includegraphics[scale = 0.4]{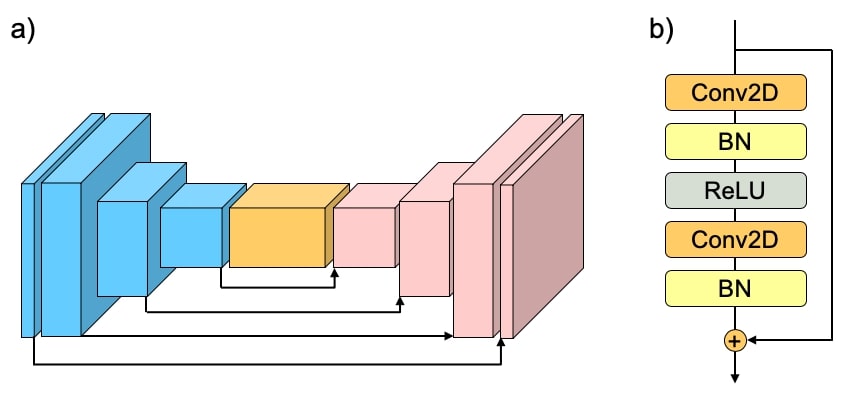}}
\caption{a) Schematic of the RU-Net. Blue represents an encoding unit, yellow represents a connecting unit, pink represents a decoding unit, and the black arrow represents a concatenating path. b) Schematic of a residual convolutional block which consists a 2D convolutional layer (Conv2D), a batch normalization layer (BN) \cite{Ioffe2015}, a rectified linear layer (ReLU) \cite{Nair2010}, a second convolutional layer, a second batch normalization layer and an identity shortcut.}
\label{network}
\end{center}
\end{figure}

Based on the U-Net architecture, in our model, the encoding path and decoding path each contains four encoder/decoder units as illustrated in Figure \ref{network}a. The latent representations of the input and output are compressed in 128 feature maps with the size of 16x16. Generally, a larger size and a greater number of the feature maps requires more trainable parameters while providing higher training accuracy. In our study, the number and the size of the feature maps are chosen based on experiments in order to balance the trade off between training accuracy and training efficiency. To further improve the performance of the deep U-Net architecture in a multiphase flow prediction problem, we used five residual convolutional blocks in the connecting unit. The residual convolutional block performs same size convolution with the ResNet framework \cite{He2015}, where a shortcut connection path learns the identity mapping within each unit and therefore eases the training requirements (Figure \ref{network}b). This RU-Net architecture is also combined with the convolutional long short-term memory (LSTM) network in \cite{tang2019deep} for capturing dynamic subsurface flow in channelized geological models and achieved promising results. The detailed RU-Net model architecture is described in Table \ref{NN_param}. 

\begin{table}[!htp]
 \caption{RU-Net Architecture. In an encoding unit, \texttt{Conv2D} denotes a 2D convolutional layer; \texttt{c} denotes the number of channels in the next layer; \texttt{k} denotes the kernel (or filter) size; and \texttt{s} denotes the size of the stride; \texttt{BN} denotes a batch normalization layer; \texttt{ReLu} denotes a rectified linear layer. In a decoding unit, \texttt{Unpool} denotes a unpooling layer, and \texttt{Padding} denotes a reflection padding layer.}
  \centering
  \begin{tabular}{lll}
    \hline
    Unit     & Layer     & Output Shape \\
    \hline
    Input    &                                   & (128,128,3) \\
    Encode 1 &  \texttt{Conv2D(c16k3s2)/BN/ReLu} & (64,64,16)  \\
    Encode 2 &  \texttt{Conv2D(c32k3s1)/BN/ReLu} & (64,64,32)  \\
    Encode 3 &  \texttt{Conv2D(c64k3s2)/BN/ReLu} & (32,32,64)  \\
    Encode 4 &  \texttt{Conv2D(c64k3s1)/BN/ReLu} & (32,32,64)  \\
    Encode 5 &  \texttt{Conv2D(c128k3s2)/BN/ReLu} & (16,16,128)  \\
    Encode 6 &  \texttt{Conv2D(c128k3s1)/BN/ReLu} & (16,16,128)  \\
   ResConv 1 &  \texttt{Conv2D(c128k3s1)/BN/ReLu/Conv2D(c128k3s1)/BN} &  (16,16,128)  \\
   ResConv 2 &  \texttt{Conv2D(c128k3s1)/BN/ReLu/Conv2D(c128k3s1)/BN} &  (16,16,128)  \\
   ResConv 3 &  \texttt{Conv2D(c128k3s1)/BN/ReLu/Conv2D(c128k3s1)/BN} &  (16,16,128)  \\
   ResConv 4 &  \texttt{Conv2D(c128k3s1)/BN/ReLu/Conv2D(c128k3s1)/BN} &  (16,16,128)  \\
   ResConv 5 &  \texttt{Conv2D(c128k3s1)/BN/ReLu/Conv2D(c128k3s1)/BN} &  (16,16,128)  \\
    Decode 6 &  \texttt{Unpool/Padding/Conv2D(c128k3s1)/BN/Relu} &  (16,16,128)  \\
    Decode 5 &  \texttt{Unpool/Padding/Conv2D(c128k3s2)/BN/Relu} &  (64,64,32)  \\
    Decode 4 &  \texttt{Unpool/Padding/Conv2D(c64k3s1)/BN/Relu} &  (64,64,32)  \\
    Decode 3 &  \texttt{Unpool/Padding/Conv2D(c64k3s2)/BN/Relu} &  (64,64,32)  \\
    Decode 2 &  \texttt{Unpool/Padding/Conv2D(c32k3s1)/BN/Relu} &  (64,64,32)  \\
    Decode 1 &  \texttt{Unpool/Padding/Conv2D(c16k3s2)/BN/Relu} &  (64,64,32)  \\
    Output   &  \texttt{Conv2D(c1k3s1)}                         & (128,128,1) \\
    \hline
  \end{tabular}
  \label{NN_param}
\end{table}

\subsubsection{Training procedures}

The loss function used for the training of this neural network is the Mean Square Error (MSE), defined as
\begin{equation}
L_{MSE}=\frac{1}{N}\sum^N_{i=1}||\mathbf{y}_i-\hat{\mathbf{y}}_i||^2_2,
\end{equation}
where $N$ is the number of training samples and $\mathbf{y}$ is the true CO${_2}$ saturation from the numerical simulator. Here $\hat{\mathbf{y}}$ is the CO${_2}$ saturation predicted by the neural network, described as
\begin{equation}
\hat{\mathbf{y}}_i=\mathbf{f}(\mathbf{x}_i,{\theta}) ,
\end{equation}
where ${\theta}$ is the trainable parameters in the neural network and $\mathbf{x}_i$ is the training input.

The MSE loss especially penalizes on the errors with larger absolute magnitude compared to those with smaller magnitude. This error is chosen because a higher CO$_2$ gas saturation in multiphase flow normally corresponds to a higher mobility, which requires a higher accuracy in plume prediction. On the contrary, small CO$_2$ saturations often appear at the edges of the plume, which might be artifacts caused by the numerical simulation technique. 

In this model, a total of 2,867,041 trainable parameters are updated during the training process to minimize the loss function described above. The update for each learnable weight is based on the gradient to the loss function with respect to the weight (referred as backpropagation in machine learning). A batch size of 16 is selected and the Adam optimizer\cite{Kingma2014} is used with the initial learning rate set to be 0.0001. 

During the training process, we use the root mean square error (RMSE) metric to monitor the model's convergence and evaluate the model's prediction performance. The RMSE error is calculated as 

\begin{equation}
RMSE=\sqrt{\frac{1}{N}\frac{1}{n}\sum^N_{i=1}||\mathbf{y}_i-\hat{\mathbf{y}}_i||^{2}_{2}},
\end{equation}

where $N$ is the number of samples that we are evaluating, $n$ is the number of grid cells of the data output, $\mathbf{y}$ is the true CO${_2}$ saturation from the numerical simulator, and $\hat{\mathbf{y}}$ is the CO${_2}$ saturation predicted by the neural network. The RMSE is a common metric in machine learning to measure model's convergence and a smaller RMSE indicates a more accurate prediction.

\subsection{Model prediction}
For model prediction, values of the permeability field, the injection rate field, and the injection duration field are fed into the previously trained neural network model. Taking the advantage of the fast prediction ability of a neural network, the model can produce each CO$_2$ plume output in around 0.0003 seconds, which is 6 orders of magnitude faster than the stage-of-the-art numerical simulator ECLIPSE.

To evaluate the performance of the neural network model, we randomly chose 30,400 data samples from the data set to be the test set and leaves 200,000 data samples for training. The RMSE in the test set is the primary comparison metric to evaluate the performance of a trained model and the test samples are never used in the training of any model. The train/test split ratio is roughly 7/1, which is adequate to provide a representative test result.

In addition, to obtain a more intuitive understanding of the accuracy for each prediction, we use the mean absolute error (MAE) to evaluate the average absolute error on each grid cell. The MAE can be calculated for either a single prediction or for a test set:
\begin{equation}
MAE=\frac{1}{N}\frac{1}{n}\sum^N_{i=1}||\mathbf{y_i}-\hat{\mathbf{y_i}}||_1 ,
\end{equation}
where $N$ is the number of samples that we are evaluating, $n$ is the number of grid cells of the CO${_2}$ saturation prediction, $\mathbf{y}$ is the true CO${_2}$ saturation output from the numerical simulator, $\hat{\mathbf{y}}$ is the CO${_2}$ saturation predicted by the neural network.

\section{Results}
In this section, we first present and analyze the test set CO$_2$ saturation results, which are predicted by the neural network model that is trained with the full data set of 200,000 samples. We then discuss the relationship between the model's performance and the training size through a sensitivity study that compares the test set results from  scenarios with different training set sizes.

\subsection{Full data set training results}

\begin{figure}[!htp]
\begin{center}
\centerline{\includegraphics[scale=0.3]{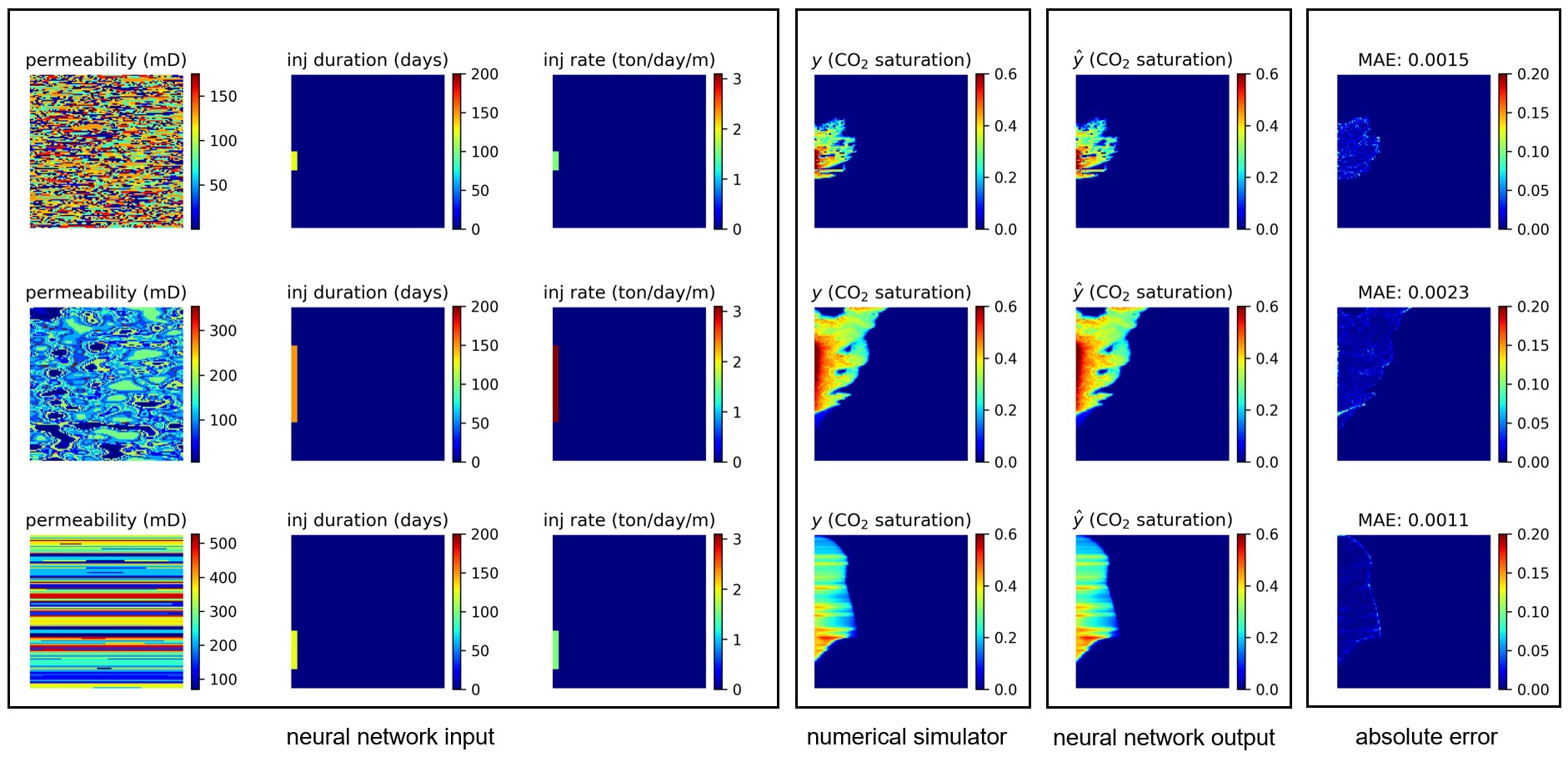}}
\caption{Test set results on the model that is trained with 200,000 samples for 50 epochs. The three examples are randomly chosen from the test set to demonstrate the model's performance for different combinations of inputs. The first to the third columns shows the input fields for each prediction. The forth and fifth column show the CO$_2$ saturation plume simulated by ECLIPSE and predicted by the neural network respectively. The sixth column shows the absolute error on each grid and the MAE is displayed on the title.}
\label{best_case}
\end{center}
\end{figure}

Using the proposed neural network architecture, loss function, and training procedure as discussed in Section 3,  we show three examples of the CO$_2$ saturation predictions in the test set (Figure \ref{best_case}). In this case, the neural network model is trained with training 200,000 samples. With this large data set, the neural network yields very accurate results for the CO$_2$ saturation maps. In the test set, the MAE is approximately 0.001, which indicates the absolute error of  CO$_2$ saturation prediction on each grid is 0.1\%. In the context of multiphase flow simulation and CO$_2$ plume migration, this error can be considered nearly negligible. 

From these examples, we can observe that the trained neural network makes the prediction of CO$_2$ plume migration according to not only the permeability field, but also the injection duration, injection rate, and well perforation. This verifies the effectiveness of using channels to incorporate variables with spatial information. Moreover, these examples demonstrate that the neural network has excellent performance with stochastic heterogeneity (first row), channelized heterogeneity (second row), and layered heterogeneity (third row). The errors in the CO$_2$ saturation prediction mostly occur at the edges of the CO$_2$ plumes where the true saturation is low. This effect is due to the choice of the loss function, which especially penalizes errors with large values.

\begin{figure}[!htp]
\begin{center}
\centerline{\includegraphics[scale=0.365]{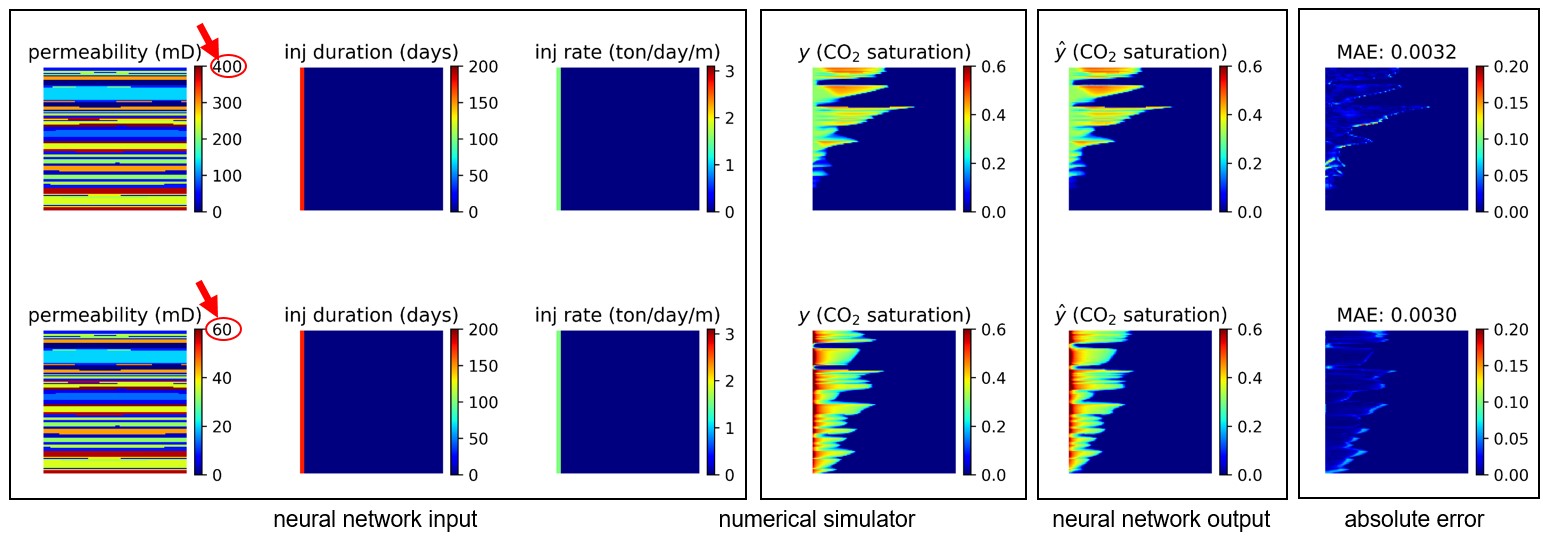}}
\caption{This example demonstrates that the network can accurately deal with the effect of capillary barriers. The input of the first row and second row are identical except that the permeability in the second row is half of the permeability in the first row.}
\label{capillary}
\end{center}
\end{figure}

More importantly, we observe that the neural network can accurately capture the complex interplay of viscous, capillary, and gravity forces. In CO$_2$ storage problems, the migration of a plume is often affected by not only the degree of permeability heterogeneity, but also the absolute value of permeability \cite{gege2019}. In Figure \ref{capillary}, we show an example of where the model needs to make prediction for two different permeability maps that have the same degree of heterogeneity but different absolute permeability. Due to the effect of capillary barriers created by the lower permeability layers in the second case, the plume on the second row is separated into multiple smaller plumes and more CO$_2$ is retained at the bottom of the reservoir. Results from Figure \ref{capillary} shows that the neural network accurately captures this behavior very well.

\begin{figure}[!htp]
\begin{center}
\centerline{\includegraphics[scale=0.3]{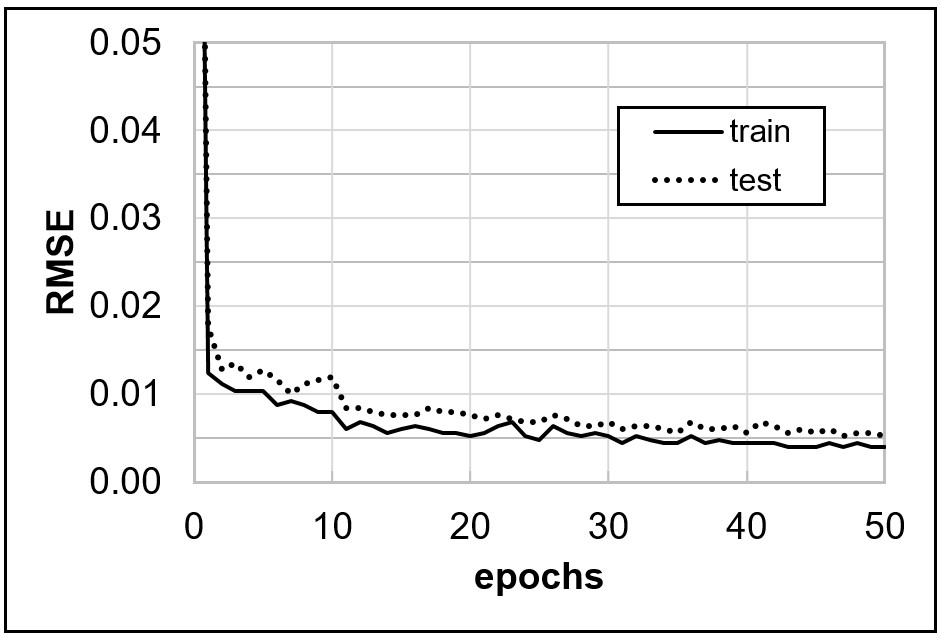}}
\caption{The evolution of the root mean absolute error (RMSE) in the training and test set to 50 epochs.}
\label{best_case_loss}
\end{center}
\end{figure}

The model discussed here was trained for 50 epochs. In deep learning, an epoch is defined as a single pass of the entire training dateset through the neural network. For the full data set described above, each epoch takes 20 to 30 minutes on an NVIDIA Tesla V100 GPU. In figure \ref{best_case_loss}, we show the evolution of the RMSE for both the training and the test set with the number of epochs. Comparing the final RMSE of the test set to the training set, we observe that the neural network model slightly overfits to the training set. In neural network training, a small degree of overfitting shows that the model performs slightly better on seen data compared to unseen data, which indicates a healthy training procedure.  

\subsection{Sensitivity to the number of training samples}
\begin{figure}[!htp]
\begin{center}
\centerline{\includegraphics[scale = 0.3]{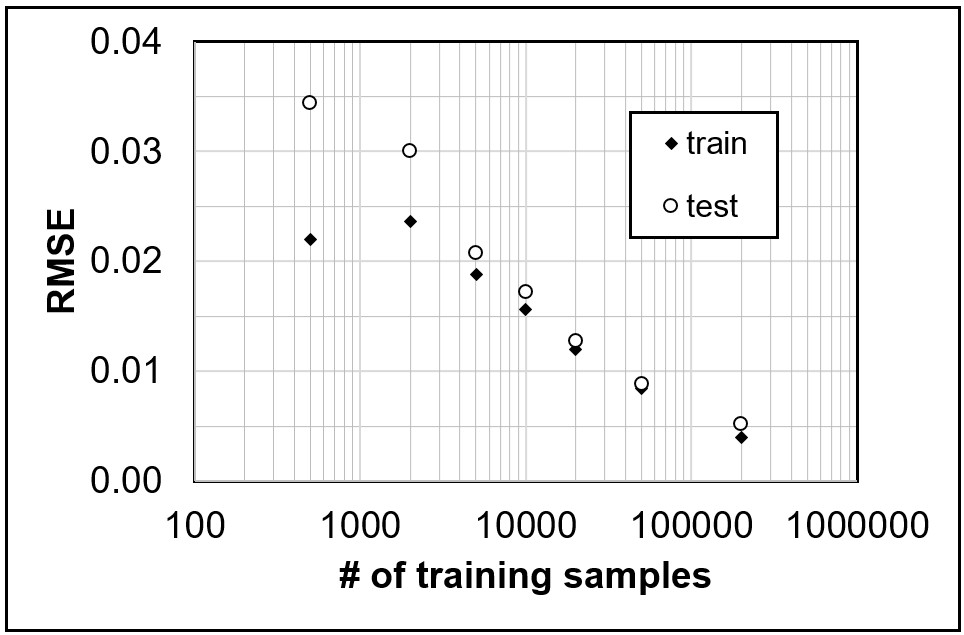}}
\caption{RMSE on the test set and training set at the end of 50 epochs verse the number of training samples.}
\label{MAE_num}
\end{center}
\end{figure}

In practice, it is often infeasible to obtain a training data set as big as the one discussed in the previous section. Therefore, to study the influence of the training data set size on the model's performance, we conduct a sensitivity study on the number of training samples where the size of the training set varies from 200,000 to 500 samples. The relationship between the number of the training samples with the test and training set RMSE at the end of 50 epochs are plotted in Figure \ref{MAE_num}. 

In general, for the test set, increasing the number of the training samples monotonically reduce the RMSE. Notice that the relationship between the RMSE and number of training samples shows a log-linear behavior, which indicates that reducing the RMSE on the test set becomes more and more difficult when the error is small as it requires an exponentially larger number of samples in the training set. We also observe that the difference between the test set RMSE and the training set RMSE increases as the number of training samples decreases, which indicates that training with smaller data sets causes more overfitting.

For the training set, the RMSE decreases as the training set size increases with the exception when the training sample size becomes too small. We can observe that when the training sample size is  500, the model achieves a lower RMSE compare to when the training sample size is 2000. This is due to an effect of severe overfitting: at a training size of only 500, the data is not sufficient to support the model to learn the full physics of this problem. Instead, the model decides to `memorize' the seen training samples, which causes a small training RMSE and a large test RMSE.

\begin{figure}[!htp]
\begin{center}
\centerline{\includegraphics[scale = 0.8]{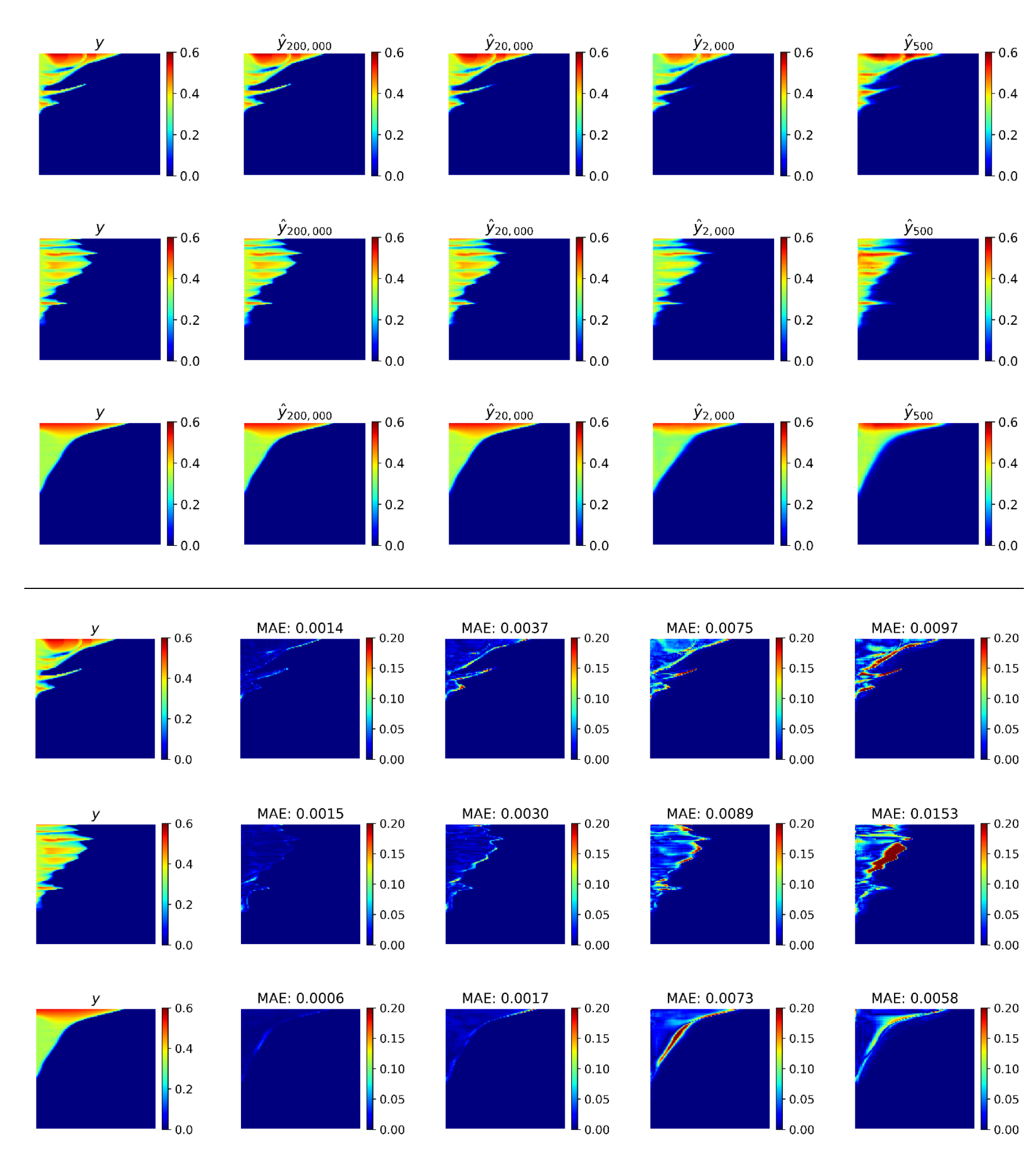}}
\caption{Row 1-3: CO$_2$ saturation plume predictions in the test set from the model trained with different sizes of the training samples. Row 4-6: Absolute error visualizations on the test set from the model trained with sizes of the of training samples. }
\label{n_comparison}
\end{center}
\end{figure}

To further investigate the neural network's performance with different sizes of the training set, three randomly picked test set predictions are shown in the first three rows of Figure \ref{n_comparison}. Predictions are compared for 200,000 down to 500 training samples. From the three examples shown in Figure \ref{n_comparison}, we can notice that the plume always migrates upward towards the top of the formation, indicating that the neural network can easily learn the effect of gravity even with the smallest training set.  Also, the predicted CO$_2$ always migrate from the well perforation given in the input of the injection field. These results indicates that the technique of incorporating the injection controlling parameters in a separate channel is effective even with a very small number of the training samples. 

In the forth to sixth rows of Figure \ref{n_comparison}, we show the comparisons of the absolute error for each of these cases. We find that that most of the improvement from using a larger training set occurs at the plume edges where the CO$_2$ saturation is low. This effect is caused by the choice of loss function which especially penalizes larger errors.

\section{Discussion}

In Section 4, we demonstrate that the neural network can make accurate predictions of the CO$_2$ saturation maps given the input of the permeability and injection field. Sensitivity studies on the number of training samples shows that the neural network can utilize efficiently the training set and achieve low training and test error even with very small data sets. However, a key challenge in predicting multiphase flow with neural networks is whether the neural network can generalize outside of its training data. In this section, we address this concern with the following generalization studies and a demonstration of a transfer learning (also referred as fine-tuning) procedure.

\subsection{Generalization}

To investigate the generalization ability of the neural network, here we studied 3 cases where the model needs to interpolate on the injection duration, extrapolate on the injection duration, and interpolate on the injection location (interval of the well perforation). For the purpose of demonstration, we use a fixed injection rate of 300 ton/day in the following interpolation and extrapolation studies and reduce the number of channels in the input to two channels. The size of the training set is 20,000 for each of the three generalization cases. In Section 3.2, the sensitivity study on the RMSE versus the training size provides a reference for when the training size is 20,000 at the end of 50 epochs, which is 0.012 for the training set and 0.013 for the test set.  


\begin{figure}[!htp]
\begin{center}
\centerline{\includegraphics[scale = 0.3]{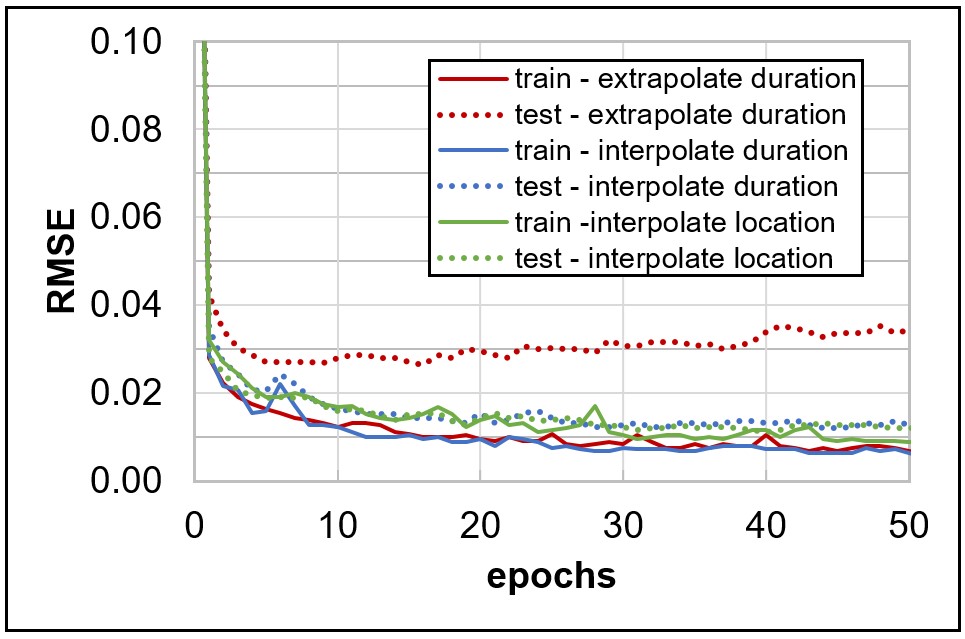}}
\caption{The test and training set RMSE evolution verses the number of epochs. The injection duration interpolation, injection duration extrapolation, and injection location interpolation cases are all plotted on this figure.}
\label{interpolate}
\end{center}
\end{figure}
 
\begin{figure}[!htp]
\begin{center}
\centerline{\includegraphics[scale = 0.32]{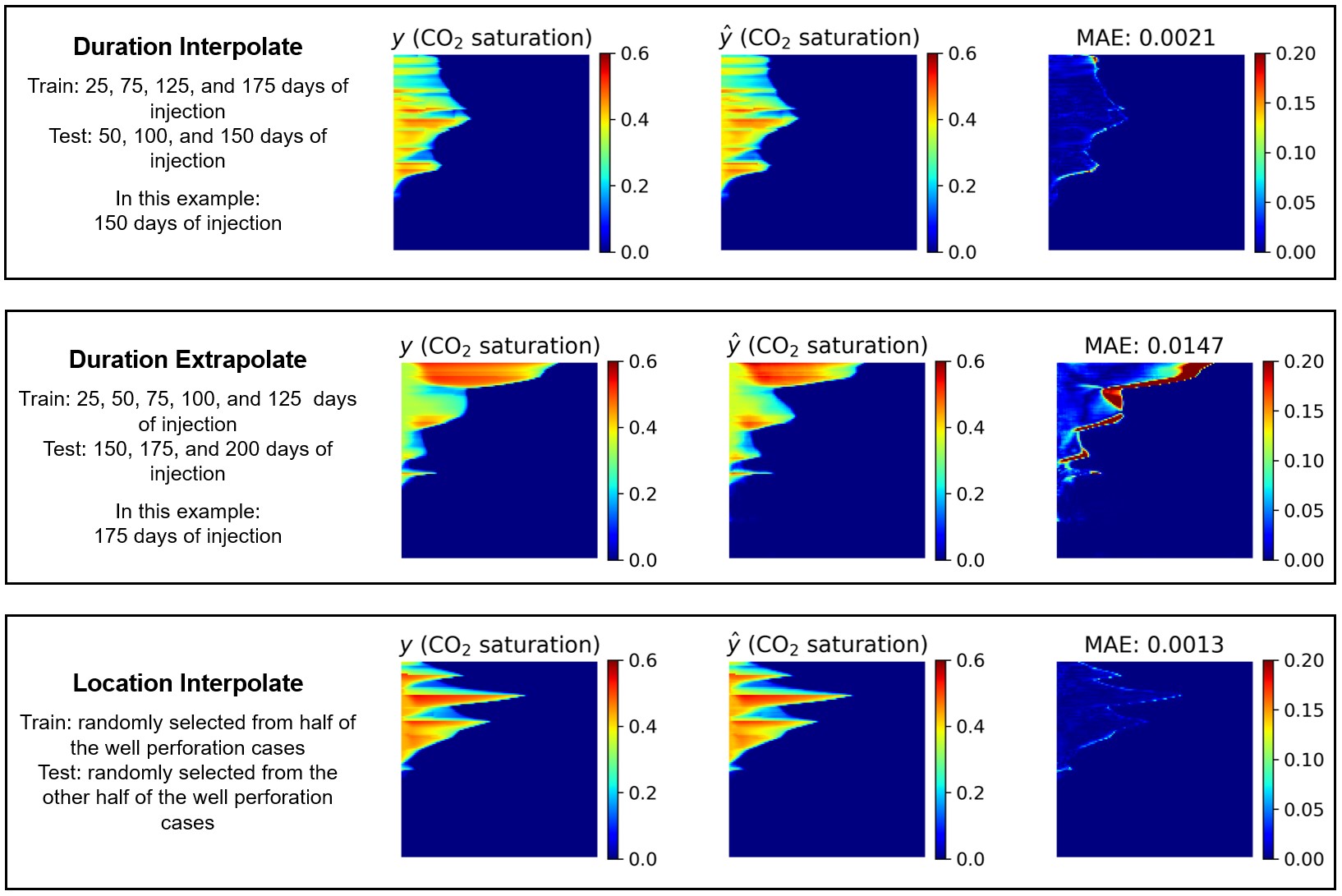}}
\caption{The first row shows an example of the test set result where the network interpolates on the injection duration. The second row shows an example of the test set result where the network extrapolates on the injection duration. The third row shows an example of the test set result where the network interpolates on the injection location.}
\label{interpolate_vis}
\end{center}
\end{figure}
 
Figure \ref{interpolate} shows the evolution of the RMSE in the training and test set with an increasing number of epochs. For the interpolation ability of the injection duration, we show a case where the training set includes samples with 20, 60, 100, 140, and 180 days of injection, and a test set that includes samples with 40, 80, 120, 160 and 200 days of injection. Similarly, the extrapolation ability is demonstrated by a training set with 20 to 120 days of injection, and a test set with 120 to 200 days of injection. The training set RMSE in both interpolation and the extrapolation cases are similar, reaches to around 0.007 after 50 epochs. Notice that the training RMSE for these two cases are lower than reference RMSE from the sensitivity study (0.012) due to a smaller data variability in the training data sets. 

For the test sets, the RMSE for the interpolation case is about 0.014 (~2 times higher than for the training set). An example from the test set is shown in the first row of Figure \ref{interpolate_vis}, in which we can observe a relatively accurate prediction with some errors on the edges of the plume. However, as for the extrapolation case, we observe an significantly different behavior of the RMSE in the test set. Not only is the extrapolation case RMSE significantly higher than the training set at the end of 50 epochs (~5 times higher than for the training set), it also shows an increasing trend with the number of epochs. This indicates a significant overfitting to the training data, where the network is more and more likely to predict a smaller CO$_2$ plume that are similar to the training data. The second row of Figure \ref{interpolate_vis} shows an example from the test set where the the neural network predicts a significantly smaller plume and yields an MAE of 1.6\%. In the context of CO$_{2}$ plume migration, this error might or might not be considered acceptable depending on the use case. To further increase the prediction accuracy when the model needs to extrapolate, methods including transfer learning (also referred as fine-tuning) with a small number of data samples can be used. Details on the transfer learning technique are discussed in the next Section.

As for the injection location, the interpolation ability is demonstrated by using a training set that contains data with 15 randomly selected well perforation scenarios, and a test set from the rest of the data set. The difference on the RMSE between the training and test set is smaller comparing to when the model interpolates on injection duration. The RMSE for the training and the test sets after 50 epochs are 0.009 and 0.012 respectively. This indicates that the model has excellent generalization ability on spatial information due to the design of the RU-Net architecture, which can effectively utilize the hierarchical spatial information of the training data. An example from the test set is shown in the third row of Figure \ref{interpolate_vis}.

\subsection{Fine tuning with site specific data}
In machine learning, when the training and test data come from different task domains, transfer learning can serve as a powerful tool which can greatly improve the performance without having to go through the massive data collection process \cite{QiangYang2010} \cite{Radenovic2018}. In the context of CO$_2$ plume migration, although the process of multiphase flow always satisfies the mass and energy balances equations, discrepancy between the training and test data can occur due to specific site properties such as injection pattern, formation depth, or formation temperature. Therefore, to transfer the knowledge that a neural network has previous learned from a large data set, we can apply a fine tuning procedure that uses the pre-trained model as the parameter initialization, and then conduct training with only a few data samples from the new task domain to broaden the applicability of the model. 

\begin{figure}[!htp]
\begin{center}
\centerline{\includegraphics[scale = 0.35]{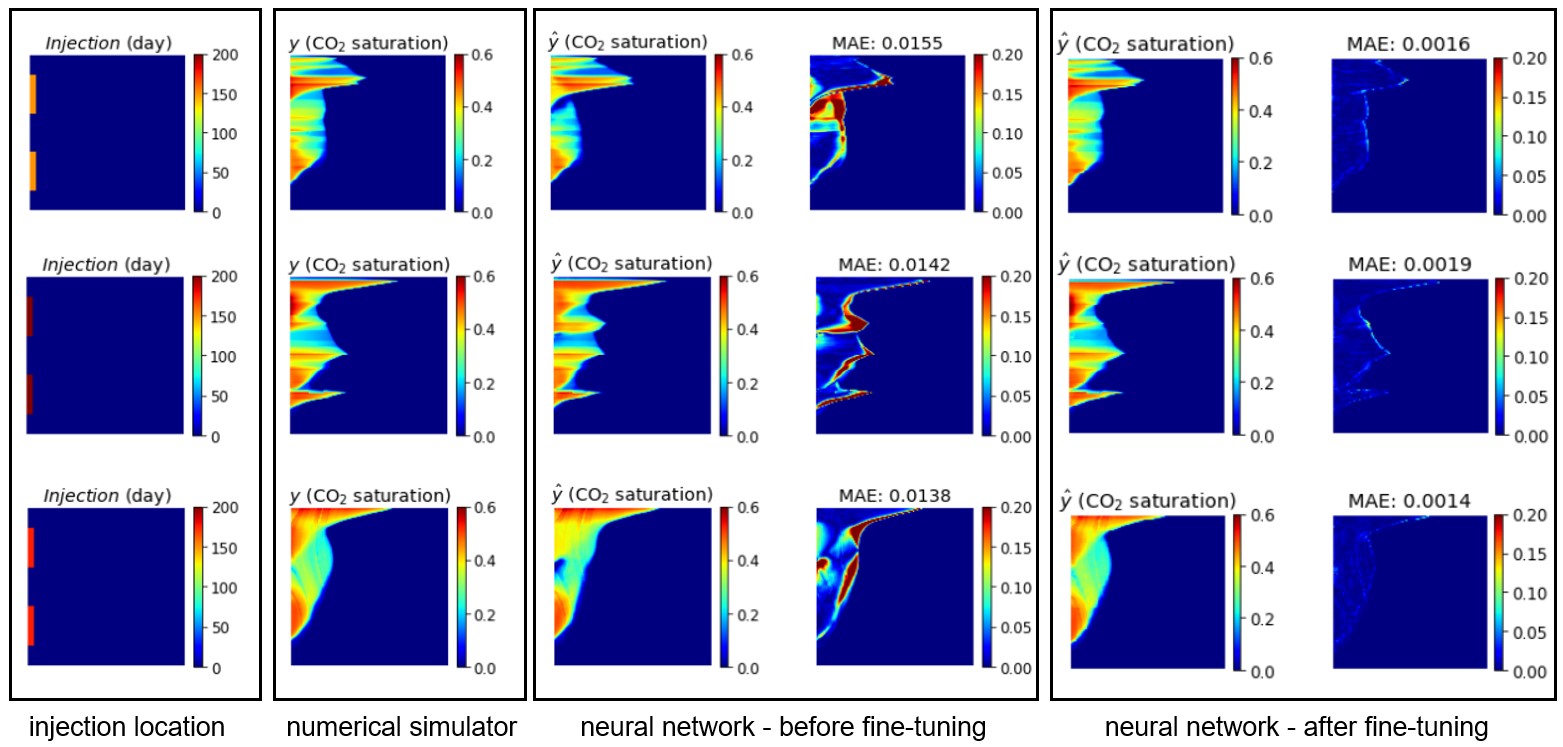}}
\caption{Example of neural network fine-tuning with data that has multiple injection well perforations. The first column shows the the location and duration of the injection. The injection rate of the three examples are 300 ton/day. The second column is the true saturation field simulated by the numerical simulator. The third and forth column are the prediction and the absolute error by the network before applying fine tuning. The fifth and the sixth column are the prediction and the absolute error by the network after applying fine-tuning.}
\label{fine-tuning}
\end{center}
\end{figure}

Here we demonstrate the feasibility of this approach by using the following example. The network that we previously discussed is trained with samples where all of the injection wells have a single continuous perforation. In reality, we sometimes see injection well with multiple sets of well perforations to satisfy operational needs. With multiple well perforation intervals, the injected CO$_2$ will form multiple plumes that may or may not merge with each other. If the lower CO$_2$ plume gets connected to the top plume, gravity will cause the CO$_2$ to migrate upward and increase the CO$_2$ saturation of the upper plume. The first column of Figure \ref{fine-tuning} shows three examples of the injection wells with multiple perforation intervals, and the second column of Figure \ref{fine-tuning} shows the numerical simulation results correspond to each perforation scenario. The injection rate of these three examples are all fixed at 300 tons/day.

First, we simply feed the multiple perforation interval input into a previously trained neural network that have never seen any input data of this type. Here we use the neural network that was trained with 200,000 training samples with a single perforation interval. This gives us the prediction and the error visualization that are shown in the third and the forth columns of Figure \ref{fine-tuning}. We can observe that this neural network has a poor prediction accuracy on CO$_2$ saturation when the CO$_2$ plumes injected from the two perforation intervals merge together. The MAE on the examples presented here increased to around 1.4\%, which is one order of magnitude larger than the MAE that we have previously achieved on the test set with a single well perforation interval.

Then, to `teach' the neural network model how the injected CO$_2$ plumes behave with multiple perforation intervals, we generated a small training set that includes 80 training samples with multiple perforation intervals. Using the per-trained model as initialization, the neural network was trained (fine-tuned) for 20 epochs. The fifth and sixth column shows that the model performs very well after fine-tuning and the MAE drops back to around 0.15\%. 

This demonstration case deals with a relatively straight forward training-test data domain shift, however, it reveals the potential of using this fine-tuning technique to broaden the applicability of a pre-trained neural network model without going through massive data collection.

\subsection{Comparison with other neural network algorithms}

In this section, we compare the performance of our RU-Net with the DenseED network used in Mo et al. \cite{Mo2019}. The comparison is performed based on the $50 \times 50$ permeability to CO$_2$ gas saturation data set that Mo et al. provides in their paper. The coefficient of determination (R$^2$) score is used for this comparison, defined as

\begin{equation}
    R^2 = 1-\frac{\sum^N_{i=1}||\mathbf{y}^i-\hat{\mathbf{y}}^i||^2_2}{\sum^N_{i=1}||\mathbf{y}^i-\bar{\mathbf{y}}^i||^2_2},
\end{equation}
where N is the number of samples in the test set; $\mathbf{y}^i$ is the true CO$_2$ saturation field; $\hat{\mathbf{y}}^i$ is the CO$_2$ saturation predicted by neural networks; and $\bar{\mathbf{y}}^i$ is the average of the true CO$_2$ saturation field in the test set. Table \ref{comparison_mo} summarizes the R$^2$ score with different training sizes.
\begin{table}[!htp]
 \caption{R$^2$ score comparison with \cite{Mo2019} with different training sizes. All results were trained for 200 epochs. MSE denotes a neural network that is trained by the mean square error; MSE-BCE denotes a neural network that is trained by a combination of mean square error and binary cross entropy error.}
  \centering
  \begin{tabular}{lllll}
    \hline
    Number of training samples     & 400    & 800  & 1200    & 1600\\
    \hline
    \cite{Mo2019} MSE    &   0.869 & 0.943 & 0.965 & 0.973  \\
    \cite{Mo2019} MSE-BCE &  0.913 & 0.953 & 0.970 & 0.976  \\
     \hline    
    (this paper) RU-Net  & 0.923 &  0.971 & 0.977 &  0.977 \\
    \hline
  \end{tabular}
  \label{comparison_mo}
\end{table}

This comparison shows that our RU-Net achieves higher R$^2$ score at all training sizes after the same amount of epochs (200) as in Mo et al.'s study. We can observe that the RU-Net is especially efficient at utilizing the data set and achieves good performance when the data set is small. In this data set, the duration of injection is a variable, which is incorporated by a second channel in our training of the RU-Net. Since this data set has a constant injection rate, the channel which represents injection rate field is simply taken out and the network can automatically adjust itself to incorporate the difference in input size. In addition, notice that their data set lies on the $xy$ plane (no gravity) while our data set lies on the $rz$ plane (with gravity). Despite the differences in the physical environment, we can still robustly apply the RU-Net to Mo et al's data set without complicated parameter tuning. This comparison confirms the excellent performance of the RU-Net that we propose, as well as the robustness of using this neural network on different multiphase flow problems.


\section{Conclusion}

In this paper, we propose a neural network based approach for making fast predictions of CO$_2$ plume migration and saturation distributions in heterogeneous reservoirs. The neural network is trained on simulation outputs from a full physics multi-phase flow simulator that includes the effects of buoyancy, capillary, and viscous forces in 2D radial settings. After setting up the training data set and model, the network can provide accurate predictions of the simulation output, in our case, the CO$_2$ saturation field. A novel and effective method is proposed to incorporate controlling parameters (the duration and rate of CO$_2$ injection) with spatial information embedded.

To show the potential of adopting this approach to real life cases, we also demonstrate the ability of this network to generalize outside of the training data. We quantitatively demonstrate that the neural network performs well on interpolating for injection duration, rate, and location compared to extrapolating for controlling variables. To tackle this problem, we introduce a fine-tuning technique which can improve the model performance on unseen data. With the tools provided in this paper, we believe that the neural network approach can be an effective and computationally efficient substitute for the repetitive forward simulations required in history-matching tasks and uncertainty analysis problems.

\acknowledgments
We acknowledge the support from the Global Climate and Energy Project at Stanford University. We acknowledge the Center for Computational Earth \& Environmental Sciences (CEES) at Stanford University for the computational resources. Upon publication of this manuscript, we will release a web-based tool that allows user to conduct CO$_2$ multiphase flow simulation online. The web tool is embedded with the pre-trained model as discussed in this manuscript. The user can upload a permeability map, select a injection rate, a injection duration, and a injection well perforation, and obtain the CO$_2$ saturation prediction result in a few seconds. The Python codes of the neural network and the full data set will also be released at https://github.com/gegewen/RU-Net upon the publication of this manuscript.

%
%

\bibliography{main}

%
%
%
%
%

\end{document}